# Integrating "Free" Word Order Syntax and Information Structure

**Beryl Hoffman***
Dept. of Computer and Information Science
University of Pennsylvania
(hoffman@linc.cis.upenn.edu)

## Abstract

Multiset-CCG is a combinatory categorial formalism that can capture the syntax and interpretation of "free" word order in languages such as Turkish. The formalism compositionally derives the predicate-argument structure and the information structure (e.g. topic, focus) of a sentence, and uniformly handles word order variation among arguments and adjuncts within a clause, as well as in complex clauses and across clause boundaries.

## 1 Introduction

In this paper, I present a categorial formalism, Multiset CCG (based on Combinatory Categorial Grammars (Steedman, 1985; Steedman, 1991)), that captures the syntax and context-dependent interpretation of "free" word order in languages such as Turkish. Word order variation in relatively free word order languages, such as Czech, Finnish, German, Japanese, Korean, Turkish, is used to convey distinctions in meaning that go beyond traditional propositional semantics. The word order in these languages serves to structure the information being conveyed to the hearer, e.g. by indicating what is the *topic* and the *focus* of the sentence (as will be defined in the next section). In fixed word order languages such as English, these are indicated largely through intonation and stress rather than word order.

The context-appropriate use of "free" word order is of considerable importance in developing practical applications in natural language generation, machine translation, and machine-assisted translation. I have implemented a database query system in Prolog, described in (Hoffman, 1994), which uses Multiset CCG to interpret and generate Turkish sentences with context-appropriate word orders. Here, I concentrate on further developing the formalism, especially to handle complex sentences.

There have been other formalisms that integrate information structure into the grammar for "free" word order languages, e.g. (Sgall et al, 1986; Engdahl/Vallduvi, 1994; Steinberger, 1994). However, I believe my approach is the first to tackle complex sentences with embedded information structures and discontinuous constituents. Multiset CCG can handle free word order among arguments and adjuncts in all clauses, as well as word order variation across clause boundaries, i.e. long distance scrambling. The advantage of using a combinatory categorial formalism is that it provides a compositional and flexible surface structure, which allows syntactic constituents to easily correspond with information structure units. A novel characteristic of this approach is that the context-appropriate use of word order is captured by compositionally building the *predicate-argument structure* (AS) and the *information structure* (IS) of a sentence in parallel.

After presenting the motivating Turkish data in Section 2, I present a competence grammar for Turkish in Section 3 that captures the basic syntactic and semantic relationships between predicates and their arguments or adjuncts while allowing "free" word order. This grammar, which derives the predicate-argument structure is then integrated with the information structure in Section 4. In Section 5, the formalism is extended to account for complex sentences and long distance scrambling.

## 2 The Turkish Data

The arguments of a verb in Turkish (as well as many other "free" word order languages) do not have to occur in a fixed word order. For instance, all six permutations of the transitive sentence below are possible, since case-marking, rather than word order, serves to differentiate the arguments.[1]

*I would like to thank Mark Steedman, Ellen Prince, and the support of NSF Grant SBR 8920230.

[1] The accusative, dative, genitive, ablative, and locative cases are associated with specific morphemes

(1) a. Fatma Ahmet'i gördü.
Fatma Ahmet-Acc see-Past.
"Fatma saw Ahmet."
b. Ahmet'i Fatma gördü.
c. Fatma gördü Ahmet'i.
d. Ahmet'i gördü Fatma.
e. Gördü Ahmet'i Fatma.
f. Gördü Fatma Ahmet'i.

Although all the permutations have the same propositional interpretation, *see(Fatma,Ahmet)*, each word order conveys a different discourse meaning only appropriate to a specific discourse situation. We can generally associate the sentence-initial position with the *topic*, the immediately preverbal position with the *focus* which receives the primary stress in the sentence, and postverbal positions with backgrounded information (Erguvanli, 1984). The post-verbal positions are influenced by the given/new status of entities within the discourse; postverbal elements are always evoked discourse entities or are inferrable from entities already evoked in the previous discourse, and thus, help to ground the sentence in the current context.

I define topic and focus according to their informational status. A sentence can be divided into a topic and a comment, where the topic is the main element that the sentence is about, and the comment is the main information we want to convey about this topic. Assuming the hearer's discourse model or knowledge store is organized by topics, the sentence topic can be seen as specifying an "address" in the hearer's knowledge store (Reinhart, 1982; Vallduvi, 1990). The informational focus is the most information-bearing constituent in the sentence, (Vallduvi, 1990); it is the new or important information in the sentence (within the comment), and receives prosodic prominence in speech. These information structure components are successful in describing the context-appropriate answer to database queries. In this domain, the focus is the new or important part of the answer to a wh-question, while the topic is the main entity that the question and answer are both about, that can be paraphrased using the clause "As for X". In other domains, finding the topic and focus of sentences according to the context may be more complicated.

We can now explain why certain word orders are appropriate or inappropriate in a certain context, in this case database queries. For example, a speaker may use the SOV order in (2b) to answer the wh-question in (2a) because the speaker wants to *focus* the new object, Ahmet, and so places it in the immediately preverbal position. However, given a different wh-question in (3), the subject, Fatma, is the *focus* of the answer, while Ahmet is the *topic*, a link to the previous context, and thus the OSV word order is used.[2]

(2) a. Fatma kimi gördü?
Fatma who-Acc see-Past?
"Who did Fatma see?"
b. Fatma Ahmet'i gördü. SOV
Fatma Ahmet-Acc see-Past.
"Fatma saw AHMET."

(3) a. Ahmet'i kim gördü?
Ahmet-Acc who see-Past.
"Who saw Ahmet?"
b. Ahmet'i Fatma gördü. OSV
Ahmet-Acc Fatma see-Past.
"As for Ahmet, FATMA saw him."

Adjuncts can also occur in different sentence positions in Turkish sentences depending on the context. The different positions of the sentential adjunct "yesterday" in the following sentences result in different discourse meanings, much as in English.

(4) a. Fatma Ahmet'i dün gördü.
Fatma Ahmet-Acc dün see-Past.
"Fatma saw Ahmet YESTERDAY."
b. Dün Fatma Ahmet'i gördü.
Yesterday Fatma Ahmet-Acc see-Past.
"Yesterday, Fatma saw Ahmet."
c. Fatma Ahmet'i gördü dün.
Fatma Ahmet-Acc see-Past yesterday.
"Fatma saw Ahmet, yesterday."

Clausal arguments, just like simple NP arguments, can occur anywhere in the matrix sentence as long as they are case-marked, (5)a and b. Subordinate verbs in Turkish resemble gerunds in English; they take a genitive marked subject and are case-marked like NPs, but they assign structural case to the rest of their arguments like verbs. The arguments and adjuncts within most embedded clause can occur in any word order, also seen in (5)a and b. In addition, elements from the embedded clause can occur in matrix clause positions, i.e. *long distance scrambling*, (5c). As indicated by the translations, word order variation in complex sentences also affects the interpretation.

(5) a.
Ayşe [dün Fatma'nın gittiğini] biliyor.
Ayşe [yest. Fatma-Gen go-Gerund-Acc] knows.
"Ayşe knows that yesterday, FATMA left."
b.
[Dün gittiğini Fatma'nin ] Ayşe biliyor.
[Yest. go-Gerund-Acc Fatma-Gen] Ayşe knows.
"It's AYŞE who knows that she, Fatma, left YESTERDAY."
c.
Fatma'nın Ayşe [dün gittiğini] biliyor.
Fatma-Gen Ayşe [yest. go-Ger-Acc] knows.
"As for Fatma, Ayşe knows that she left YESTERDAY."

(and their vowel-harmony variants) which attach to the noun; nominative case and subject-verb agreement for third person singular are unmarked.

[2] In the English translations, the words in capitals indicate phonological focus.

The information structure (IS) is distinct from predicate-argument structure (AS) in languages such as Turkish because adjuncts and elements long distance scrambled from embedded clauses can take part in the IS of the matrix sentence without taking part in the AS of the matrix sentence.

As motivated from the data, a formalism for "free" word order languages such as Turkish must be flexible enough to handle word order variation among the arguments and the adjuncts in all clauses, as well as the long distance scrambling of elements from embedded clauses. In addition, to capture the context-appropriate use of word order, the formalism must associate information structure components such as topic and focus with the appropriate sentence positions, regardless of the predicate-argument structure of the sentence, and be able to handle the information structure of complex sentences. In the next sections I will present a combinatory categorial formalism which can handle these characteristics of "free" word order languages.

## 3 "Free" Word Order Syntax

In Multiset-CCG[3], we capture the syntax of free argument order within a clause by relaxing the subcategorization requirements of a verb so that it does not specify the linear order of its arguments. Each verb is assigned a function category in the lexicon which subcategorizes for a *multiset* of arguments, without linear order restrictions. For instance, a transitive verb has the category $\boldsymbol{S|\{Nn, Na\}}$, a function looking for a set of arguments, a nominative case noun phrase ($Nn$) and an accusative case noun phrase ($Na$), and resulting in the category $S$, a complete sentence, once it has found these arguments in any order.

The syntactic category for verbs provides no hierarchical or precedence information. However, it is associated with a propositional interpretation that does express the hierarchical ranking of the arguments. For example, the verb "see" is assigned the lexical category $\boldsymbol{S : see(X, Y)|\{Nn : X, Na : Y\}}$, and the noun "Fatma" is assigned $\boldsymbol{Nn : Fatma}$, where the semantic interpretation is separated from the syntactic representation by a colon. These categories are a shorthand for the many syntactic and semantic features associated with each lexical item. The verbal functions can also specify a *direction* feature for each of their arguments, notated in the rules as an arrow above the argument. Thus, verb-final languages such as Korean can be modeled by using this direction feature in verbal categories, e.g. $S|\{\overleftarrow{Nn}, \overleftarrow{Na}\}$.

Multiset-CCG contains a small set of rules that combine these categories into larger constituents. The following application rules allow a function such as a verbal category to combine with one of its arguments to its right (>) or left (<). We assume that a category $X|\emptyset$ where there are no arguments left in the multiset rewrites by a clean-up rule to just $X$.

(6) a. **Forward Application (>):**

$$X|(Args \cup \{\overrightarrow{Y}\}) \quad Y \;\Rightarrow\; X|Args$$

b. **Backward Application (<):**

$$Y \quad X|(Args \cup \{\overleftarrow{Y}\}) \;\Rightarrow\; X|Args$$

Using these application rules, a verb can apply to its arguments in any order. For example, the following is a derivation of a transitive sentence with the word order Object-Subject-Verb; variables in the semantic interpretations are italicized.[4]

(7)

| | | |
|---|---|---|
| Ahmet'i | Fatma | gördü. |
| Ahmet-Acc | Fatma | saw. |
| Na:Ahmet | Nn:Fatma | S: see($X$, $Y$)\|{Nn:$X$,Na:$Y$} |

————————————————<

S:see(Fatma, $Y$)| {Na:$Y$}

————————————————–<

S: see(Fatma, Ahmet)

In fact, all six permutations of this sentence can be derived by the Multiset-CCG rules, and all are assigned the same propositional interpretation, *see(Fatma,Ahmet)*.

The following composition rules combine two functions with set-valued arguments, e.g. two verbal categories, a verbal category and an adjunct.

(8) a. **Forward Composition (>B):**

$$X|\,(\mathrm{Args}_X \cup \{\overrightarrow{Y}\}) \quad Y|\,\mathrm{Args}_Y \;\Rightarrow\; X|\,(\mathrm{Args}_X \cup \mathrm{Args}_Y)$$

b. **Backward Composition (<B):**

$$Y|\,\mathrm{Args}_Y \quad X|\,(\mathrm{Args}_X \cup \{\overleftarrow{Y}\}) \;\Rightarrow\; X|\,(\mathrm{Args}_X \cup \mathrm{Args}_Y)$$

c. Restriction: $Y \neq NP$.

Through the use of the composition rules, Multiset-CCGs can handle the free word order of sentential adjuncts. Adjuncts are assigned a function category $S|\{S\}$ that can combine with any function that will also result in $S$, a complete sentence. The same composition rules allow two verbs to compose together to handle complex sentences with embedded clauses. This will be discussed further in section 5.

The restriction $Y \neq NP$ on the Multiset-CCG composition rules prevents the categories for verbs, $S|\{NP\}$, and for adjectives, $NP|\{\overrightarrow{NP}\}$, from combining together before combining with a bare noun. This captures the fact that simple NPs must be continuous and head-final in Turkish. Multiset CCG is flexible enough to handle

[3] A preliminary version of the syntactic component of the grammar was presented in (Hoffman, 1992).

[4] In my implementation of this grammar, DAG-unification is used in the rules. To improve the efficiency of unification and parsing, the arguments of the categories represented as DAGS are associated with feature labels that indicate their category and case.

"free" word order languages that are freer than Turkish, such as Warlpiri, through the use of unrestricted composition rules, but it can also handle languages more restrictive in word order such such as Korean by restricting the categories that can take part in the composition rules.

## 4 The Discourse Meaning of "Free" Word Order

Word order variation in Turkish and other "free" word order languages is used to express the information structure of a sentence. The grammar presented in the last section determines the predicate-argument structure of a sentence, regardless of word order. In this section, I add the ordering component of the grammar where the information structure of a sentence is determined. The simple compositional interface described below allows the AS and the IS of a sentence to be derived in parallel. This interface is very similar to Steedman's approach in integrating prosody and syntax in CCGs for English (Steedman, 1991).

A. Each Multiset-CCG category encoding syntactic and semantic properties in the AS is associated with an Ordering Category which encodes the ordering of IS components.

B. Two constituents can combine if and only if

   i. their syntactic/semantic categories can combine using the Multiset-CCG application and composition rules,

   ii. and their Ordering Categories can combine using the rules below:
   **Simple Forward Application (>):** $X/Y \quad Y \Rightarrow X$.
   **Simple Backward Application (<):** $Y \quad X\backslash Y \Rightarrow X$.
   **Identity (=):** $X \quad X \Rightarrow X$

Every verbal category in Multiset-CCG is associated with an ordering category, which serves as a template for the IS. For example, the ordering category in (9) is a function that specifies the components which must be found to complete a possible IS. The forward and backward slashes in the category indicate the direction in which the arguments must be found, and the parentheses around arguments indicate optionality. The variables $T, F, G1, G2$ will be unified with the interpretations of the proper constituents in the sentence during the derivation.

(9)
$I/$ (Ground: $G2$)$\backslash$Topic: $T\backslash$ (Ground: $G1$)$\backslash$Focus: $F$
where $I =$

$$\left[\begin{array}{ll} \text{Topic :} & T \\ \text{Comment :} & \left[\begin{array}{ll} \text{Focus :} & F \\ \text{Ground :} & [\text{verb}, G1, G2\,] \end{array}\right] \end{array}\right]$$

The function above can use the simple application rules to first combine with a focused constituent on its left, then a ground constituent on its left, then a topic constituent on its left, and a ground constituent on its right. This function will result in a complete IS only if it finds the obligatory sentence-initial topic and the immediately preverbal focus constituent; its other arguments (the ground) are optional and can be skipped during the derivation through a category rewriting rule, $X|(Y) \Rightarrow X$, that may apply after the application rules.[5]

Nonverbal elements are associated with simpler ordering categories, often just a variable which can unify with the topic, focus, or any other component in the IS template during the derivation. The identity rule allows two constituents with the same discourse function (often variables) to combine. These simpler ordering categories also contain a feature which indicates whether they represent given or new information in the discourse model, which is dynamically checked during the derivation. Restrictions (such that elements to the right of the verb have to be discourse-old information in Turkish) are expressed as features on the arguments of the verbal ordering functions.

What is novel about this formalism is that the predicate-argument structure and the information structure of a sentence are built in parallel in a compositional way. For example, given the following question, we may answer in a word order which indicates that "today" is the topic of the sentence, and "Little Ahmet" is the focus. The derivation for this answer is seen in Figure 1.

(10) a. Bugün kimi görecek Fatma?
Today who-Acc see-Fut Fatma?
"As for today, who will Fatma see?"

b. Bugün küçük Ahmet'i görecek Fatma.
Today little Ahmet-Acc see-Fut Fatma.
"Today, she, Fatma, will see Little AHMET."

In Figure 1, every word in the sentence is associated with a lexical category right below it, which is then associated with an ordering category in the next line. Parallel lines indicate the application of rules to combine two constituents together; the first line is for combining the syntactic categories, and the second line is for combining the ordering categories of the two constituents. The syntactic constituents are allowed to combine to form a larger constituent, only if their pragmatic counterparts (the ordering categories) can also combine. Thus, the derivation reflects the single surface structure for the sentence, while compositionally building the AS and the IS of the sentence in

[5] Another IS is available where the topic component is marked as "inferrable", for those cases where the topic is a zero pronoun instead of an element which is realized in the sentence. After the derivation is complete, further discourse processing infers the identity of the unrealized topic from among the salient entities in the discourse model.

(11)

| Bugün | Küçük | Ahmet'i | gördü | Fatma. |
|---|---|---|---|---|
| Today | little | Ahmet-Acc | saw | Fatma. |
| S:today($P$)\|{S:$P$} | N$x$:little($Z$)/N$x$:$Z$ | Na:Ahmet | S: see($X,Y$)\|{Nn:$X$, Na:$Y$} | Nn:Fatma |
| $X$:today | $Y$:little | $Z$:Ahmet | $I$/(Grnd2)\Top\(Grnd1)\Foc | $W_{given:+}$:Fatma |

————————————>
————————————=
AS = Na:little(Ahmet)
IS = $Y$: [little,Ahmet]
————————————————————<
————————————————————-<,skip
AS = S:see($X$,little(Ahmet)) | { Nn:$X$}
IS = [Focus:[little,Ahmet],Ground:see]/(Grnd2)\ Top
————————————————————————————————>B
————————————————————————————————-<
AS = S: today(see($X$,little(Ahmet))) | { Nn:$X$}
IS = [Topic: today, Focus:[little,Ahmet], Ground:see]/(Grnd2)
————————————————————————————————->
————————————————————————————————->
AS = S: today(see(Fatma, little(Ahmet)))
IS = [Topic: today, Focus: [little,Ahmet], Ground: [see,Fatma]]

Figure 1: Deriving the Predicate-Argument and Information Structure for a Simple Sentence.

parallel.

Using this formalism, I have implemented a database query system (Hoffman, 1994) which generates Turkish sentences with context-appropriate word orders, in answer to database queries. In generation, the same topic found in the database query is maintained in the answer. For wh-questions, the information that is retrieved from the database to answer the question becomes the focus of the answer. I have extended the system to also handle yes-no questions involving the question morpheme "mi", which is placed next to whatever element is being questioned in the sentence. If the verb is being questioned, this is a cue that the assertion or negation of the verb will be the focus of the answer:

(12) a. Ahmet'i Fatma gördü mü?
Ahmet-Acc Fatma see-Past Quest.
"As for Ahmet, did Fatma SEE him?"

b. Hayır, Ahmet'i$_T$ Fatma [GÖRmedi]$_F$.
No, Ahmet-Acc Fatma see-Neg-Past.
"No, (as for Ahmet) Fatma did NOT see him."

In most Turkish sentences, the immediately preverbal position is prosodically prominent, and this corresponds with the informational focus. However, verbs can be focused in Turkish by placing the primary stress of the sentence on the verb instead of immediately preverbal position and by lexical cues such as the placement of the question morpheme. Thus, we must have more than one IS available for verbs, where verbs can be in the focus or the ground component of the IS. In addition, it is possible to focus the whole VP or the whole sentence, which can be determined by the context, in this case the database query:

(13) a. Bugün Fatma ne yapacak?
Today Fatma what do-Fut?
"What's Fatma going to do today?"

b.
Bugün Fatma [kitap okuyacak]$_F$.
Today Fatma book read-fut.
"Today, Fatma is going to [read a BOOK]$_F$

In yes/no questions, if a non-verbal element is being focused by the question morpheme and the answer is no, the system provides a more natural and helpful answer by replacing the focus of the question with a variable and searching the database for an alternate entity that satisfies the rest of the question.

Thus, Multiset CCG allows certain pragmatic distinctions to influence the syntactic construction of the sentence using a lexicalized compositional method. In addition, it provides a uniform approach to handle word order variation among arguments and adjuncts, and as we will see in the next section, across clause boundaries.

# 5 Complex Sentences

## 5.1 Embedded Information Structures

As in matrix clauses, arguments and adjuncts in *embedded* clauses can occur in any order. To capture the interpretation of the word order within embedded clauses, my formalism allows for embedded information structures. Subordinate verbs, just like matrix verbs, are associated with an ordering category which determines the information structure for the clause. When the subordinate clause syntactically combines with the matrix clause, the IS of the subordinate clause is embedded into the IS of the matrix clause. For example, in the complex sentence and its IS below, the embedded clause is the topic of the matrix clause since it occurs in the sentence-initial position of the matrix clause. The word order variation within the embedded clause indicates the structure of the IS that is embedded under topic.

(14) a. [Dün Fatma'nin gittiğini] Ayşe biliyor.
[Yest. Fatma-Gen go-Ger-Acc] Ayşe knows.
"It's AYŞE who knows that yesterday, FATMA left."

$$\begin{bmatrix} \text{Topic :} & \begin{bmatrix} \text{Topic :} & \text{yesterday} \\ \text{Comment :} & \begin{bmatrix} \text{Focus :} & \text{Fatma} \\ \text{Ground :} & \text{go} \end{bmatrix} \end{bmatrix} \\ \text{Comment :} & \begin{bmatrix} \text{Focus :} & \text{Ayşe} \\ \text{Ground :} & \text{know} \end{bmatrix} \end{bmatrix}$$

To ensure that the embedded IS is complete before it is placed into the matrix clause's IS, we restrict the application rules (e.g. $X/Y \;\; Y \Rightarrow X$) in the ordering component of Multiset-CCG; we stipulate that the argument Y must not be a function (with arguments left to find). The restriction ensures that the ordering category for the embedded verb is no longer a function, that it has found all of its obligatory components and skipped all the optional ones before combining with the matrix verb's ordering category.

### 5.2 Long Distance Scrambling

In Turkish complex sentences with clausal arguments, elements of the embedded clauses can occur in matrix clause positions, i.e. long distance scrambling. However, speakers only use long distance scrambling for specific pragmatic functions. Generally, an element from the embedded clause can occur in the sentence initial topic position of the matrix clause (e.g. (15)b) or to the right of the matrix verb as backgrounded information (e.g. (15)d), but cannot occur in the stressed immediately preverbal position (e.g. (15)c). This long distance dependency is similar to the English topicalization construction.

(15) a. Ayşe [Fatma'nin dün gittiğini] biliyor.
Ayşe [Fatma-Gen yesterday go-Ger-Acc] knows.
"Ayşe knows that Fatma left yesterday."

b. Fatma'nın Ayşe [dün gittiğini] biliyor.
Fatma-Gen Ayşe [yest. go-Ger-Acc] knows.

c. *Ayşe [dün gittiğini] FATMA'nın biliyor.
*Ayşe [yest. go-Ger-Acc] Fatma-Gen knows.

d. Ayşe [dün gittiğini] biliyor Fatma'nın.
Ayşe [yest. go-Ger-Acc] knows Fatma-Gen.

Multiset-CCG can recover the appropriate predicate-argument relations of the embedded clause and the matrix clause even when the arguments occur out of the domain of the subordinate verb. The composition rules allow two verb categories with multisets of arguments to combine together. As the two verbs combine, their arguments collapse into one argument set in the syntactic representation. As seen in the derivation below, we compose the verbs together to form a complex verbal function, which can then apply to the arguments of both verbs in any order.

(16)

| | |
|---|---|
| gittiğini | biliyor |
| go-gerund-acc | knows |
| $S_{Na} : go(y) \vert \{Ng{:}y\}$ | $S{:}know(x,p) \vert \{Nn{:}\ x, S_{na}{:}\ p\}$ |

——————————————————<B

$S : know(x, go(y)) \vert \ \{Nn : x,\ Ng : y\}$

Although the verbs' argument sets are collapsed into one set, their respective arguments are still distinct within the semantic representation of the sentence. The propositional interpretation of the subordinate clause is embedded into the interpretation of the matrix clause.

The syntactic component of Multiset-CCGs correctly rules out long distance scrambling to the immediately preverbal matrix position, because elements from the embedded clause cannot combine with the matrix verb before the matrix verb has combined with the embedded verb.

(17)

| | | | |
|---|---|---|---|
| *[Gittiğini] | Ayşe | Fatma'nin | biliyor. |
| *[Go-Ger-Acc] | Ayşe | Fatma-Gen | know-Pres. |
| $S_{Na} \vert \{Ng, Na\}$ | Nn | Ng | $S \vert \{Nn, S_{Na}\}$ |

——————XXX——————

Long distance scrambling to the sentence initial position and post-verbal position in the matrix clause is handled through the composition of the verbs, as seen in Figure 2.

The ordering component of Multiset CCG allows individual elements from subordinate clauses to be components in the IS of the matrix clause. This is because the ordering category for a matrix verb does not specify that its components be arguments in its AS. In the sentence in Figure 2, "Fatma", an argument of the embedded clause, has been scrambled into the topic position of the matrix clause. The derivation with both components of the grammar working in parallel is shown in Figure 2. The embedded verb must first complete its IS ($IS_2$); then, the two verbs compose together, and the subordinate IS is embedded into the matrix IS ($IS_1$). The complex verbal constituent can then combine with the rest of the arguments of both verbs in any order. The linear order of the two NP arguments will determine which components of the matrix IS each fill. Note that "Fatma" is an argument in the interpretation of the embedded verb "go", not the matrix verb "know", but it plays the role of topic in the matrix

| Fatma'nin | Ayşe | [dün | gittiğini] | biliyor. |
|---|---|---|---|---|
| Fatma-Gen | Ayşe | [yesterday | go-Ger-Acc] | know-Pres. |
| Ng:Fatma | Nn:Ayşe | S:yest($P$)\|{S:$P$} | $S_{Na}$: go($X$) \| { Ng:$X$ } | S: know($Y$,$Z$) \| { Nn:$Y$, $S_{Na}$ : $Z$} |
| $X$:Fatma | $Y$:Ayşe | $W$:yesterday | $IS_2$/(Grnd2)\(Top)\(Grnd1)\Foc | $IS_1$/(Grnd2)\Top\(Grnd1)\Foc |

——————————————————>B

——————————————————<,skip3

AS = $S_{Na}$: yesterday(go($X$)) | { Ng:$X$}

$IS_2$ = [Topic:inferrable, Focus:yesterday, Ground:go]

——————————————————————————————-<B

——————————————————————————————<

AS = S: know($Y$,yesterday(go($X$))) | { Nn:$Y$, Ng:$X$}

$$IS_1 = \left[\begin{array}{ll} \text{Topic :} & Top \\ \text{Comment :} & \left[\begin{array}{ll} \text{Focus :} & IS_2 \left[\begin{array}{ll} \text{Topic :} & \text{inferrable} \\ \text{Comment :} & \left[\begin{array}{ll} \text{Focus :} & \text{yest.} \\ \text{Ground :} & \text{go} \end{array}\right] \end{array}\right] \\ \text{Ground :} & [\text{know}, Grnd1, Grnd2] \end{array}\right] \end{array}\right] /(\text{Grnd2})\backslash\text{Top}\backslash(\text{Grnd1})$$

————————————————————————————————<

————————————————————————————————<

AS = S: know(Ayşe,yesterday(go($X$))) | { Ng:$X$}

IS = $IS_1/(Grnd2)\backslash Top$

————————————————————————————————<

————————————————————————————————<

AS = S: know(Ayşe,yesterday(go(Fatma)))

$$IS = \left[\begin{array}{ll} \text{Topic :} & \text{Fatma} \\ \text{Comment :} & \left[\begin{array}{ll} \text{Focus :} & \left[\begin{array}{ll} \text{Topic :} & \text{inferrable} \\ \text{Comment :} & \left[\begin{array}{ll} \text{Focus :} & \text{yesterday} \\ \text{Ground :} & \text{go} \end{array}\right] \end{array}\right] \\ \text{Ground :} & [\text{Ayşe, know}] \end{array}\right] \end{array}\right]$$

Figure 2: Derivation for the AS and IS of a Complex Sentence.

verb's IS. Thus, adjuncts and elements from embedded clauses can play a role in the information structure of the matrix clause, although they do not belong to the same predicate-argument structure.

### 5.3 Islands

The syntactic component of Multiset-CCGs can derive a string of any number of scrambled NPs followed by a string of verbs: $(NP_1 ... NP_m)_{scrambled}\ V_m\ ...\ V_1$, where each verb, $V_i$, subcategorizes for $NP_i$. The more one scrambles things, the harder the sentence is to process, but there is no clear cut-off point in which the scrambled sentences become ungrammatical for native speakers. Thus, I claim that processing limitations and pragmatic purposes, rather than syntactic competence, restrict such scrambling.

However, some types of clauses, in some "free" word order languages, act as islands that strictly do not allow long distance scrambling. In other "free" word order languages, such as Turkish, it is very hard to find island effects. As seen in the first example in Figure 3, even elements from relative clauses can be extracted. However, it is harder to extract elements from some adjunct clauses which do not have close semantic links to the matrix clause. To account for clauses exhibiting island behaviour, we can assign the head of the clause a category such as $S|S|\{Nn, Na\}$ which makes certain that the head combines with all of its NP arguments before combining with the matrix clause, $S$. As demonstrated in (19)c in Figure 3, long distance scrambling out of such an adjunct clause is thus prohibited.

In contrast, heads of adjunct clauses which are not islands are assigned categories such as $S|\{S, Nn, Na\}$. Since this category can combine with the matrix verb even before it has combined with all of its arguments, it allows long distance scrambling of its arguments. This lexical control of the behaviour is very advantageous for capturing Turkish, since not every adjunct clause is an island in Turkish. However, further research is

(18) *Ankara'dan*$_i$ sen [e$_i$ dün gelen] adamı tanıyor musun?
*Ankara-Abl*$_i$ you [e$_i$ yest. come-Rel] man-Acc know Quest-2Sg?
"Do you know the man who came yesterday from Ankara?"

(19) a. [Berna ödevini bitirince] bana yardım edecek.
[Berna hw-3Ps-Acc finish-ger] I-dat help do-3Sg.
"When Berna finishes (her) homework, (she) is going to help me."

b. *[Berna bitirince] bana yardım edecek *ödevini.*
*[Berna finish-ger] I-dat help do *hw-3Ps-Acc.*

c. ∗Berna finish-ger I-dat help do hw-3Ps-Acc
$Nn$ $S|S|\{Nn, Na\}$ ———$S$——— $Na$
——————————<
$S|S|\{Na\}$
———XXX——————— ———XXX———

Figure 3: Long Distance Scrambling Out of Adjunct Clauses

needed to determine what types of adjunct clauses exhibit island behaviour in order to specify the appropriate categories in the lexicon.

## 6 Conclusions

I have presented a combinatory categorial formalism that can account for both the syntax and interpretation of "free" word order in Turkish. The syntactic component of Multiset CCG is flexible enough to derive the predicate-argument structure of simple and complex sentences without relying on word order, and it is expressive enough to capture syntactic restrictions on word order in different languages such as languages with NP or clausal islands or languages which allow discontinuous NPs or clauses. Word order is used in the ordering component of Multiset CCG to determine the information structure of a sentence. Every Multiset CCG category encoding syntactic and semantic properties is associated with an ordering category which encodes the ordering of information structure components such as topic and focus; two syntactic/semantic categories are allowed to combine to form a larger constituent only if their ordering categories can also combine. The formalism has been implemented within a database query task in Quintus Prolog, to interpret and generate simple and complex sentences with context-appropriate word orders.

Multiset CCG captures the context-appropriate use of word order by compositionally deriving the predicate-argument structure and the information structure of a sentence in parallel. It allows adjuncts and elements from embedded clauses to take part in the information structure of the matrix clause, even though they do not take part in its predicate-argument structure. Thus, this formalism provides a uniform approach in capturing the syntactic and pragmatic aspects of word order variation among arguments and adjuncts, and across clause boundaries.